\documentclass[conference]{IEEEtran}
\IEEEoverridecommandlockouts
\usepackage{cite}
\usepackage{amsmath,amssymb,amsfonts}
\usepackage{algpseudocode}
\usepackage{graphicx}
\usepackage{textcomp}
\usepackage{amsthm}
\usepackage{xcolor}
\def\BibTeX{{\rm B\kern-.05em{\sc i\kern-.025em b}\kern-.08em
    T\kern-.1667em\lower.7ex\hbox{E}\kern-.125emX}}

\theoremstyle{definition}
\newtheorem{definition}{Definition}[section]

\begin{document}

\title{Super-convergence and Differential Privacy: Training faster with better privacy guarantees}

\author{
\IEEEauthorblockN{Osvald Frisk}
\IEEEauthorblockA{\textit{Department of Engineering} \\
\textit{Aarhus University, Denmark}\\
201511615@post.au.dk
}
\and
\IEEEauthorblockN{Friedrich Dörmann}
\IEEEauthorblockA{\textit{Department of Engineering} \\
\textit{Aarhus University, Denmark}\\
201911254@post.au.dk
}
\and
\IEEEauthorblockN{Christian Marius Lillelund}
\IEEEauthorblockA{\textit{Department of Engineering} \\
\textit{Aarhus University, Denmark}\\
cl@eng.au.dk
}
\and
\IEEEauthorblockN{Christian Fischer Pedersen}
\IEEEauthorblockA{\textit{Department of Engineering} \\
\textit{Aarhus University, Denmark}\\
cfp@eng.au.dk
}
}

\maketitle

\begin{abstract}
The combination of deep neural networks and Differential Privacy has been of increasing interest in recent years, as it offers important data protection guarantees to the individuals of the training datasets used. However, using Differential Privacy in the training of neural networks comes with a set of shortcomings, like a decrease in validation accuracy and a significant increase in the use of resources and time in training. In this paper, we examine super-convergence as a way of greatly increasing training speed of differentially private neural networks, addressing the shortcoming of high training time and resource use. Super-convergence allows for acceleration in network training using very high learning rates, and has been shown to achieve models with high utility in orders of magnitude less training iterations than conventional ways. Experiments in this paper show that this order-of-magnitude speedup can also be seen when combining it with Differential Privacy, allowing for higher validation accuracies in much fewer training iterations compared to non-private, non-super convergent baseline models. Furthermore, super-convergence is shown to improve the privacy guarantees of private models.
\end{abstract}


\section{Introduction}
Privacy-preserving data analysis is becoming increasingly important as technologies for curating and collecting data grow in their capacities at an ever increasing rate. To preserve privacy of individuals in a dataset, Differential Privacy (DP), a mathematically rigorous definition of privacy presented by Dwork et al. \cite{Dwork_DP_14} has in recent years become the standard in addressing privacy in data analysis and especially machine learning.

However, while DP offers various, obvious benefits to machine learning in terms of privacy guarantees to the individuals of a dataset, it also comes with a set of shortcomings. These include that the training process consumes many more computational resources and takes a much longer time, furthermore limiting the possible model architectures. Additionally, the accuracy and therefore utility of differentially private models is usually lower than that of non-private counterparts.
So as of now when training neural networks, a deep learning practitioner would have to choose between a fast training, better performing model with a more richer selection of possible architectures which does not deliver any privacy guarantees and one that offers privacy guarantees but performs significantly worse in the other previously mentioned areas.
Usually, this difference in performance is so large, that non-private models have to be chosen purely for practical reasons. A goal in the research of differentially private training of neural networks is therefore close this gap between private and non-private training.

Consequently, with this work we present a way that allows for much faster training of differentially private deep learning models, effectively eliminating the mentioned shortcomings to a large degree. This is achieved through super-convergence, a novel approach to learning rate scheduling, first developed by Smith et al. \cite{Smith_SC_17}. Super-convergence has been shown to allow the training of neural networks orders of magnitude faster than other common practices.

\subsection{Context and related research project}
Since much of research is based on collecting data from a population, the recent increase in privacy concerns and resulting governmental regulations can hinder research in its effectiveness and speed. Differential Privacy addresses these issues by promising to solve problems regarding privacy for both the data contributor and the data consumer. For the data contributor, i.e. individuals, it offers a strong guarantee that their privacy is preserved and for the data consumer, i.e. scientists conducting research, it offers a tool for easier access to data, while eliminating concerns about leaking information of individuals.

One of the many scientific efforts using Differential Privacy for the aforementioned benefits is the \emph{Artificial Intelligence in Rehabilitation (AIR)} project \cite{air_project}. The project aims to develop and apply artificial intelligence approaches to support citizens in physical rehabilitation in Danish municipalities. In the AIR project, models are trained on citizen specific and sensitive data under EU's General Data Protection Regulation (GDPR) and must be kept private at all times. The work in this paper has been conducted in connection to the AIR project, however without specific data from the project being included in the paper.

\section{Differential Privacy in Machine Learning}

\subsection{Differential Privacy}
The definition of Differential Privacy provides a way of determining an upper-bound for the loss of privacy of a dataset given some statistical query or function over the dataset. This upper-bound is often referenced to as the privacy guarantee and in its original form is quantified by one value, $\varepsilon$, referenced as the privacy budget \cite{Dwork_DP_14}.

\theoremstyle{definition}
\begin{definition}[$\varepsilon$-Differential Privacy \cite{Dwork_DP_06}]\label{e_DP_Definition} A randomized algorithm $\mathcal{M}$ is $\varepsilon$-differentially private if for all datasets $D_1$ and $D_2$ differing on at most one element, and all possible solutions, $S$, of $\mathcal{M}$ applied on $D_1$ or $D_2$

\[Pr [\mathcal{M}(D_1) \in S] \leq \,e ^{\varepsilon} \cdot Pr [\mathcal{M}(D_2) \in S]\]
\end{definition}

From the definition, it is apparent that for any adjacent datasets, the difference in probability that the result of an algorithm $\mathcal{M}$ has been reached with either one of the datasets is bounded by $e^\varepsilon$. It follows that lower values of epsilon yield better privacy guarantees, as for $\varepsilon=0$, the probabilities are equal.

Another definition given as direct consequence of definition~\ref{e_DP_Definition} is that of \emph{privacy loss}.

\theoremstyle{definition}
\begin{definition}[Privacy Loss \cite{Dwork_DP_14}]\label{PL_Definition}
\[ \mathcal{L}^{(\xi)}_{\mathcal{M}(x) || \mathcal{M}(y)} = \ln \left(\frac{Pr[\mathcal{M}(D_1) = \xi]}{Pr[\mathcal{M}(D_2) = \xi]}\right)\]
\end{definition}

The Privacy Loss incurred by observing an event $\xi$ can be positive or negative, depending on whether the event is more likely under $x$ or under $y$. It can furthermore be shown, that $\varepsilon$-Differential Privacy ensures that for all adjacent $x$ and $y$, the absolute value of the privacy loss is bounded by the privacy budget $\varepsilon$ \cite{Dwork_DP_14}.\\

\subsubsection{Influencing the privacy guarantee}
A way of influencing the privacy guarantee of a statistical query is by the addition of randomly sampled noise (Laplacian, Gaussian, etc.). This noise is parameterized by the uniqueness of the data in the dataset with the query applied on it. Worded differently, for some queries one might not need to add any noise, since the query might not disclose any information about a specific observation or individual in the dataset. For other queries on the other hand, the L1-sensitivity of the query on the dataset would need to be determined to calculate the additive noise and subsequently also the $\varepsilon$ \cite{Dwork_DP_14}.\\

\subsubsection{Composition}
A property of Differential Privacy that becomes especially important in the context of machine learning is composition. Composition in its most basic form, i.e. the basic composition theorem \cite{Dwork_DP_14}, is the property that when two algorithms with privacy budgets $\varepsilon_1$ and $\varepsilon_2$ are applied on a dataset, the total privacy guarantee amounts to the summation of $\varepsilon_1$ and $\varepsilon_2$ \cite{Dwork_DP_14} \cite{Eval_DP_19}.

\subsection{Relaxed Definitions of Differential Privacy}
Composition becomes more important, the more differentially private queries are to be performed on a dataset, since the privacy budget spent increases with every query. In the case of machine learning, where it is not uncommon that an input dataset is queried up to millions of times in training, the privacy budget can increase drastically, which in turn hurts the privacy guarantees that can be given.
Therefore, possible improvements to composition are very valuable and have been a big focus for research in recent years, as they possibly allow for a better selection of noise such that one can increase the utility of a model while spending less privacy budget. Such improved composition theorems are usually based on different and relaxed definitions of DP itself.\\

\subsubsection{($\varepsilon, \delta$)-Differential Privacy}
The first important relaxation of definition was presented by Dwork \cite{Dwork_DP_14}. In it, a newly introduced delta term encapsulates a density of probability, that the bound given by $\varepsilon$ does not hold. The extended definition can be seen in Definition \ref{ed_DP_Definition}.

\begin{definition}[($\varepsilon, \delta$)-Differential Privacy \cite{Dwork_DP_14}]\label{ed_DP_Definition} A randomized algorithm $\mathcal{M}$ is ($\varepsilon, \delta$)-differentially private if for all adjacent datasets $D_1$ and $D_2$ and all $\mathcal{S} \subseteq$ Range($\mathcal{M}$)
\[Pr [\mathcal{M}(D_1) \in S] \leq \,e ^{\varepsilon} \cdot Pr [\mathcal{M}(D_2) \in S] + \delta\]
\end{definition}

This relaxation was able to form the basis for a new composition theorem, which allowed for better utility of sets of differentially private queries on data, called the Advanced Composition Theorem or Advanced Composition (AC). AC works by evaluating the linear composition of expected privacy loss of algorithms, which is then converted to a cumulative budget $\varepsilon$ with a high probability bound. Using this advanced composition theorem, the composed privacy budget $\varepsilon$ could be reduced significantly at the cost of a slight increase of delta \cite{Eval_DP_19}.\\

\subsubsection{Concentrated Differential Privacy and Rényi Differential Privacy}
Since this first relaxation of Differential Privacy, much work has gone into finding reformulations and relaxations of the ($\varepsilon, \delta$)-DP definition to improve AC, many of which are variants of another relaxation called Concentrated Differential Privacy (CDP) \cite{CDP_16}.
These variations to the definition achieve tighter guarantees, by virtue of their analysis of cumulative privacy loss which takes into account that the privacy loss random variable follows a sub-Gaussian distribution and is therefore strictly centered around an expected value. Multiple compositions of differentially private mechanisms therefore result in the aggregation of corresponding mean and variance values of the individual sub-Gaussian distributions, which in turn can be converted to a cumulative privacy budget similar to AC. All in all, this reduces the noise that must be added to the individual mechanisms \cite{CDP_16}\cite{Eval_DP_19}.

In CDP, the metric for measuring the difference between sub-gaussian distributions is the sub-gaussian divergence. CDP uses different measures than $\varepsilon$ and $\delta$ to quantify privacy and can not be mapped back to ($\varepsilon, \delta$)-DP. However, as presented by Mironov in \cite{RDP_17}, exchanging sub-gaussian divergence in the definition by Rényi Divergence yields another definition, which supports this mapping, known as \emph{Rényi Differential Privacy (RDP)}. RDP defines Differential Privacy in terms of bounding the divergence of two Distributions of a Mechanism $\mathcal{M}$ over two adjacent datasets, using Rényi Divergence of a specific order as the measure of difference. Thus, the definition is as follows

\begin{definition}[Rényi Differential Privacy \cite{RDP_17}]\label{RDP_Definition} A randomized algorithm $\mathcal{M}$ is said to have $\varepsilon$-Rényi Differential Privacy of order $\alpha$ (($\alpha, \varepsilon$)-RDP), if for any adjacent datasets $D_1$ and $D_2$ it holds that
\[\mathcal{D}_{\alpha}(\mathcal{M}(D_1) || \mathcal{M}(D_2)) \leq \varepsilon\]
\end{definition}

As presented by Jayaraman et al. 2019 \cite{Eval_DP_19}, when comparing ($\varepsilon, \delta$)-DP and AC with three state of the art variants of CDP, namely CDP, zero-CDP and RDP, computing privacy loss using RDP provides the best performance, therefore allowing the training of more accurate models with the least amount of privacy spent when compared to the other variants. In their work, they show RDP-based methods to require an order of magnitude less privacy budget $\varepsilon$ to the next best method (zero-CDP) to achieve the same model utility. For this reason, for the remainder of this paper, RDP will serve as the primary definition of choice to analyze the composition of differentially private functions applied on training datasets.

\subsection{Differentially private training of machine learning models}
Making machine learning training differentially private requires adding noise at some point in the process. The most prominent way of adding noise in training of deep neural networks is Gradient Perturbation. This method adds noise to the gradients of the loss function at each time the gradients are passed backwards through the network to update the network parameters during training. Additionally, the gradients are clipped in L2-norm, to provide a sensitivity bound on the gradients \cite{DLDP_16}.

When adding noise in the training algorithm itself, a crucial component of training becomes the tracking of the total amount of noise added, i.e. privacy budget spent, throughout the process. Introduced by Abadi et al. \cite{DLDP_16} in 2016, one of the most commonly used methods of tracking the spent privacy for the training of a deep learning model is the \emph{moments accountant}. The moments accountant keeps track of the bound on the moments of the privacy loss random variable, utilizing that the privacy loss is itself dependent on the moments of the randomly added perturbation noise.

The accountant computes the log moments of the privacy loss random variable, which compose linearly, and uses these moments together with the standard Markov inequality to bound the tail probabilities of the privacy loss.  In its original form, this procedure was based on improvements to the strong composition theorem, based on the already mentioned AC. In its recent implementations that are incorporated in the most popular Differential Privacy libraries for deep learning, the moments accountant is used in combination with RDP-composition to analyze the spent privacy budget \cite{Opacus_20} \cite{TF_Privacy_20}. This is therefore also the method of choice for tracking the privacy budget in this paper.

\section{Shortcomings of differentially private learning}
As mentioned, Differential Privacy comes with some drawbacks when used in machine learning. This work focuses specifically on the drawbacks of decreased model utility and increased model training time.

\subsection{Decreased model utility}
One shortcoming of differentially private learning is decreased predictive performance. Differentially privately trained models are in most cases inferior to the utility of their non-differentially private counterparts.

When using the original moments accountant with gradient perturbation, most machine learning models required very large values for $\varepsilon$. In previous approaches, e.g. the model presented by Shokri Et al. \cite{Shokri_15}, the utility of the model was the priority, with a disregard for the size of $\varepsilon$. This yielded a model that needed an $\varepsilon$ comparable to the number of parameters of the deep neural network, in the order of hundreds of thousands, at which point the size of the $\varepsilon$ cripples any privacy guarantees of the model.

There has been substantial developments in differential private deep learning since then, leading to the most recent version of the moments accountants based on RDP for composition. These make it possible to reduce the total loss in privacy over many iterations of training a deep neural network, such that the upper-bounds on privacy i.e. the privacy budget is much tighter, from hundreds of thousands, or even millions down to e.g. 50 for a large multi-class classification such as CIFAR-100 or less than 10 for a smaller multi-class classification problem such as MNIST \cite{Eval_DP_19}. However, it is still the case that the utilities of differentially private models are substantially worse than their non-private counterparts.

\subsection{Increased training time}
Gradient perturbation is currently the standard for training differentially private deep neural networks and while this method of training is able to yield much better privacy guarantees than other methods, it also comes with the drawback that to obtain the best possible epsilon, gradients needs to be perturbed on a single sample basis, i.e. training the deep neural network with a batch size of 1. This dramatically increases the training time by orders of magnitude, from minutes or seconds to train a non-private model with a larger batch size to hours and days for a private variant of the same model. It also eliminates the use of GPUs in training, since they offer no direct benefit over CPUs when training with such a small batch size. A considerable amount of the work in the field of private deep learning apply dimension reduction techniques such as Principal Components Analysis (PCA) in combination with smaller network architectures to combat these increased training times \cite{Eval_DP_19} \cite{DLDP_16} \cite{Greene_talk_18}.

\section{Addressing the shortcomings}
Just like $\varepsilon$ is a composite of the privacy loss for each update of the models parameters, one might consider the utility of the model a composite of each update of the parameters. Given that a lot of effort has been put into minimizing the total composition of $\varepsilon$ across the training of these private models, this work focuses on the task of maximizing the composite utility of a privately trained model, while at the same time minimizing the training time to reach this utility. This problem can therefore be stated as a maximization problem of trying to maximize model utility contingent on minimal training, and it can be approached like a hyperparameter tuning problem.
Therefore, in this work it was decided to target learning rate scheduling as the focus of the following experiments as a means of accelerating model learning at training time. Learning rate as a hyperparameter is both independent from the gradient perturbation and thus does not affect the composite privacy loss and there have also been great advances in very high learning rate scheduling research, enabling deep neural networks to converge on a high utility solution orders of magnitude faster than the classic low and static or low and close to static learning rate - a phenomenon known as super-convergence.

\subsection{Super-Convergence}
Super-convergence is a phenomenon discovered by Leslie Smith et al. \cite{Smith_SC_17} in 2018, with which neural networks can be trained one to several orders of magnitude faster than with standard training methods.

This work presented several learning rate schedules, one of which is the one cycle learning rate schedule. In this approach, the learning rate is changed continuously throughout training, the first and the last values being very small learning rates and the middle values being very large learning rates.

While one might think that this would cause models to be more prone to overfitting, the opposite is the case. The method has been shown to work as a type of regularization, and has been demonstrated to achieve competitive validation accuracy scores on tasks like MNIST and ImageNet with a reduction in the number of training iterations by 85\% \cite{Smith_SC_17}.

\section{Experiments and Results}
This chapter shows the results of three experiments of using super-convergence in differentially private training of neural networks. The first experiment compares the utility, i.e. the validation accuracies of the trained models as a function of the number of training epochs, the second experiment compares the utility of the models as a function of the privacy guarantees $\epsilon$, and finally the third experiment shows insights in how the privacy guarantees of the trained models develops as a function of the number of training epochs.

\subsection{Experimental Setup}
Throughout the experiments, we evaluate differentially private convolutional neural networks using gradient perturbation for a non-convex learning problem. Per experiment, two CNN models are compared, one of which uses a conventional learning rate schedule of decreasing the learning rate by an order of magnitude on validation loss plateau and the other utilizing the 1cycle learning rate policy proposed by Smith et al. to achieve super-convergence \cite{Smith_DA_18}.

The experiments were conducted on the MNIST dataset for multi-class classification \cite{LeCun_MNIST_98}, consisting of 70000 28x28 images of handwritten digits. The images were split into a training and validation set of 60000 and 10000 images respectively. The experiments are implemented in Python using the PyTorch deep learning library \cite{Pytorch_20}. The models in all experiments are convolutional neural networks (CNN) with a total of five layers, consisting of two 2D convolutional layers, one 2D Dropout layer and two fully connected linear layers. The hyperparameters of the model, namely the learning rate, and the number of epochs, were trained using the Bayesian Optimization and Hyperband (BOHB)  \cite{Falkner_BOHB_18} as well as the development tool Weights and Biases, to automate and supervise the tuning process \cite{Wandb_20}.

When comparing differentially private models, the model utility is represented by their \emph{validation accuracy}. When comparing differentially private models with a non-private baseline model, it is represented as \emph{accuracy loss}. Accuracy loss normalizes the model's validation accuracy with respect to the non-private baseline ($\varepsilon$ = $\infty$) as such

\[ \text{Accuracy Loss} = 1 - \frac{\text{Accuracy of Private Model}}{\text{Accuracy of Non-Private Model}}\]

Finally, the privacy guarantees for each of the models are determined by the moments accountant method using RDP-based composition. In this regard, two different implementations were used to demonstrate technology independence, the first being the newly developed official PyTorch Differential Privacy library \emph{Opacus}, as well \emph{PyVacy}, an open source project which was the most used Differential Privacy library for PyTorch before the release of Opacus \cite{Opacus_20} \cite{Pyvacy_20}.

\subsection{Experiment 1 - Training resource use and model utility}
The first experiment demonstrates the differences in training resource usage and achieved utility between a super-convergent DP model and a conventionally DP model. Resource usage is in this case presented as the amount of epochs i.e. training iterations used in the training process, which encapsulates both computational resources and training time spent.  \\

\begin{figure}
\centerline{\includegraphics[scale=0.47]{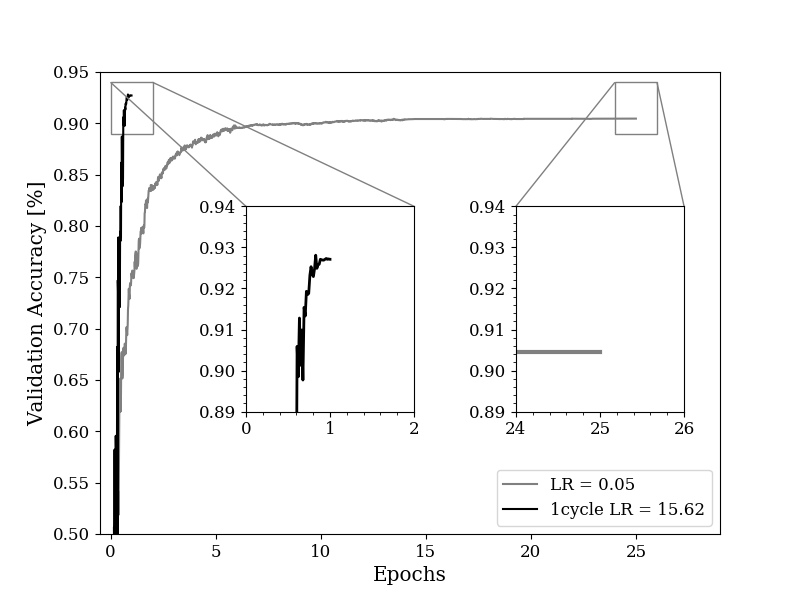}}
\caption{Validation Accuracies throughout the training process of Exp. 1}
\label{Fig_Exp1}
\end{figure}

The results can be seen in Figure \ref{Fig_Exp1} and show that the model trained with super-convergence achieves a validation accuracy of 92.7\% in only 1 epoch, while the non-superconvergent model reaches 90.5\% in 25 training epochs.

This experiment demonstrates that that using super-convergence on the differentially private CNN allows the model to converge in a fraction of the epochs of the conventional approach while yielding a higher validation accuracy. As mentioned by \cite{Smith_SC_17} \cite{Smith_DA_18}, the use of high learning rates accomplishes regularization of the model, therefore adding to the regularization by gradient perturbation, that the model is exposed to to begin with. This results in a model that is on par and even better performing on the validation set, which is visible in the improved accuracy.

\begin{figure}
\centerline{\includegraphics[scale=0.47]{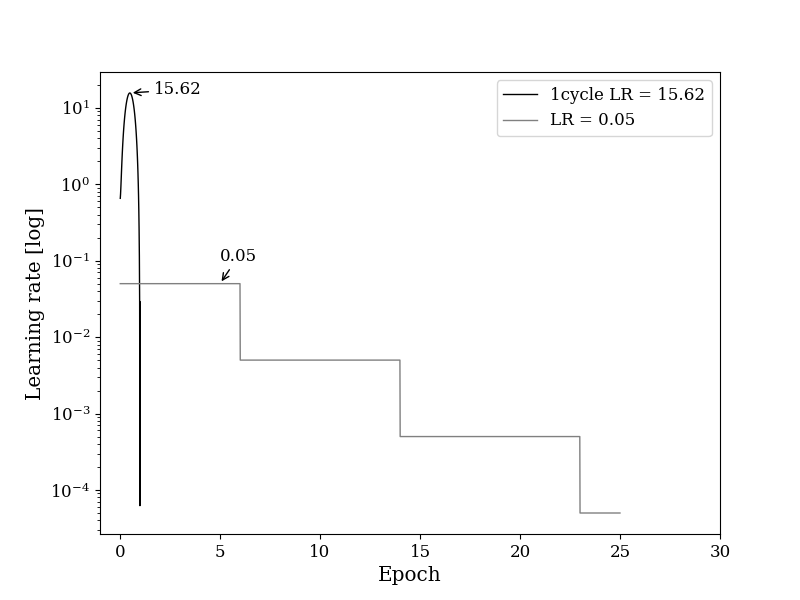}}
\caption{Learning rates throughout the training process of Exp. 1}
\label{Fig_Exp2}
\end{figure}

The super-convergent model in this case was trained with a 1cycle policy using a maximum learning rate of 15.62, while the non-superconvergent model used a learning of 0.05 at the start, which decreased by three orders of magnitude throughout the training process. The learning rate schedules are displayed in Figure \ref{Fig_Exp2}.

\subsection{Experiment 2 - Privacy guarantees and model utility}
The second experiment evaluates the models utilities against the privacy guarantees. Again, a CNN using the aforementioned plateau based learning rate schedule is compared to a CNN using the 1cycle learning rate super-convergence approach. In this case, utility is measured as accuracy loss against a non-private baseline CNN of the same architecture, which achieved an accuracy of 99.1\%.

The \emph{privacy guarantee} is parameterized by ($\varepsilon, \delta$)-Differential Privacy, however $\delta$ is held constant at $10^{-5}$, which is considered standard practice \cite{Dwork_DP_14} for differentially private machine learning with a dataset of the given size of MNIST. This leaves $\varepsilon$ as the variable to quantify the difference between the compared models.\\

\begin{figure}
\centerline{\includegraphics[scale=0.47]{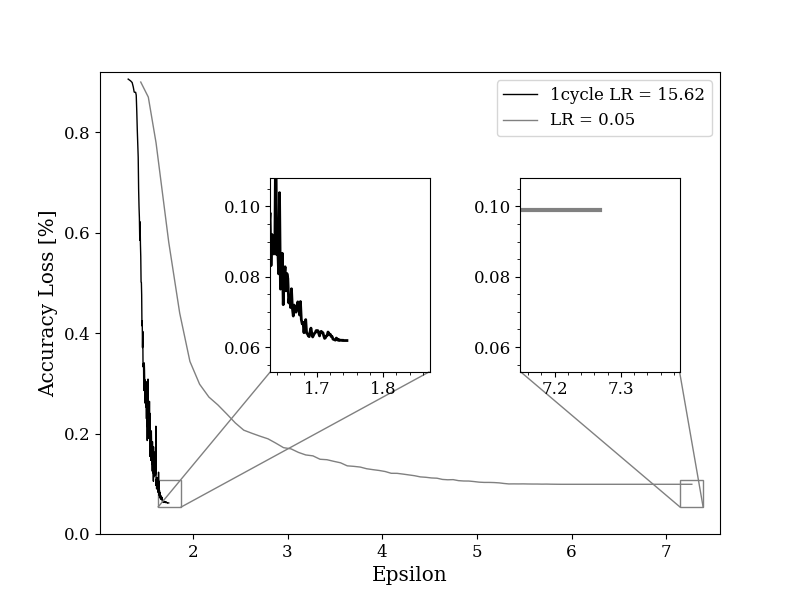}}
\caption{Accuracy Loss as a function of privacy guarantees in Exp. 2}
\label{Fig_Exp3}
\end{figure}

The results can be seen in Figure \ref{Fig_Exp3}. The super-convergent model exhibits an accuracy loss of 6.3\% with a privacy budget of $\varepsilon = 1.75$ spent during training. The non-super-convergent model shows an accuracy loss of 9.8\% compared to the baseline, with the spent privacy budget at $\varepsilon = 7.27$.

None of the two differentially private CNN’s accomplished as high a validation accuracy as the baseline, as can be seen by none of the models achieving an accuracy loss of zero. However, with accuracy loss in both cases being below 0.1, the models obtain almost similar accuracies and around that of the study Abadi et al. \cite{DLDP_16}.
The major difference between the models is the privacy guarantee expressed in epsilon, with the model trained using super-convergence able to achieve a lower accuracy loss while also having a more than four times lower epsilon, yielding better privacy guarantees.

\subsection{Experiment 3 - Privacy guarantees and training epochs}
The previous two experiments demonstrated the utility of the CNN models as a function of epochs and privacy guarantees, respectively. The third and final experiment examines the epsilon as a function of the number of training epochs for two differentially private CNN models. \\

\begin{figure}
\centerline{\includegraphics[scale=0.47]{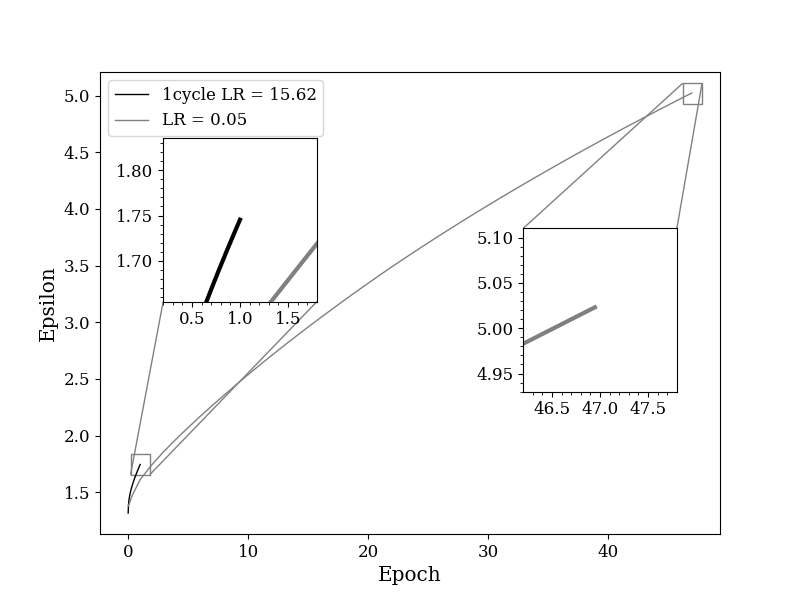}}
\caption{Privacy guarantees throughout the training process in Exp. 3}
\label{Fig_Exp4}
\end{figure}

The results can be seen in Figure \ref{Fig_Exp4}. The two models, that reached very similar, comparable validation accuracies with 93\% and 92\% for the super-convergent CNN and the regular CNN respectively, show privacy budget spent of $\varepsilon = 1.75$ and $\varepsilon = 5.025$ respectively.

The results show that in this case of two models with virtually same utility, the privacy guarantee of the superconvergent model is about three times as strong after training. As is visible in the plot, this is can be mostly attributed to the short training duration, which is an order of magnitude lower in terms of the amount of epochs.

\section{Discussion}
In summary, the experiments show strong results in regards to using super-convergence to accelerate private training of neural networks. They demonstrate that the use of a 1cycle learning rate policy with high maximum learning rates can lead to models of the same or even higher utility, while taking considerably less time to train and also offering better privacy guarantees in the process.

One thing to note however is that, while the strength of Differential Privacy is that it offers guarantees for the upper bound of leaked private information, the utility of machine learning models usually depends on finding a very specific set of hyperparameters, which often requires many iterations of hyperparameter tuning using sophisticated search methods. Adding Differential Privacy to deep learning in turn even increases the amount of hyperparameters that have to be optimized for. So while the experiments covered in this paper do show great promise for the use of super-convergence in the private network training, it cannot be used to make guarantees in terms of which training approach is the best, but should merely be viewed as a suggestion to which parts of the hyperparameter space to explore. Especially because the theory of super-convergence is still in its early development and not yet fully understood in the scientific community.

\section{Conclusion}
In conclusion, the experimental results show that differentially private models trained using super-convergence perform just as well or better in terms of validation accuracy as the non-super-converging variants. The super-convergent models take more than an order of magnitude less time and computational resources to train while consuming only a fraction of the privacy budget. Therefore, the gap between non-private and private neural networks in terms of validation accuracy, and training resource use can be decreased significantly using this method. This improves the usability of differentially private models greatly and expands the possible model architectures of differentially private models, all while providing both better privacy guarantees and model utility.

While these experimental results are promising, the robustness of this approach when applied to other learning problems, e.g. regression or larger multi-class classification, has yet to be demonstrated on differentially private models. We therefore see these problems as candidates for future work on the topic. However, super-convergence has been demonstrated to work on regular models for these problems and we are therefore optimistic in its applicability.

\section*{Acknowledgment}
We would like to thank the AIR project \cite{air_project} for financial support.

\vspace{12pt}

\end{document}